\documentclass[runningheads]{llncs}

\usepackage{eccv}

\usepackage{eccvabbrv}

\usepackage{graphicx}
\usepackage{booktabs}

\usepackage[accsupp]{axessibility}  

\usepackage[pagebackref,breaklinks,colorlinks,citecolor=eccvblue]{hyperref}

\usepackage{orcidlink}
\usepackage{multirow}
\usepackage{makecell}
\usepackage{caption}
\usepackage{subcaption}
\usepackage{colortbl}

\begin{document}

\title{Subspace Prototype Guidance for Mitigating Class Imbalance in Point Cloud Semantic Segmentation} 

 \titlerunning{Subspace Prototype Guidance for Mitigating Class Imbalance}

\author{Jiawei Han\inst{1}\orcidlink{0000-0002-6143-3868} \and
Kaiqi Liu\inst{1}\thanks{Corresponding author, liukaiqi@bit.edu.cn} \and
Wei Li\inst{1} \and
Guangzhi Chen\inst{2}}

\authorrunning{J. Han et al.}

\institute{Beijing Institute of Technology \and Beihang University}

\maketitle
 
\begin{abstract}
	 Point cloud semantic segmentation can significantly enhance the perception of an intelligent agent. Nevertheless, the discriminative capability of the segmentation network is influenced by the quantity of samples available for different categories. To mitigate the cognitive bias induced by class imbalance, this paper introduces a novel method, namely subspace prototype guidance (\textbf{SPG}), to guide the training of segmentation network. Specifically, the point cloud is initially separated into independent point sets by category to provide initial conditions for the generation of feature subspaces. The auxiliary branch which consists of an encoder and a projection head maps these point sets into separate feature subspaces. Subsequently, the feature prototypes which are extracted from the current separate subspaces and then combined with prototypes of historical subspaces guide the feature space of main branch to enhance the discriminability of features of minority categories. The prototypes derived from the feature space of main branch are also employed to guide the training of the auxiliary branch, forming a supervisory loop to maintain consistent convergence of the entire network. The experiments conducted on the large public benchmarks (i.e. S3DIS, ScanNet v2, ScanNet200, Toronto-3D) and collected real-world data illustrate that the proposed method significantly improves the segmentation performance and surpasses the state-of-the-art method. The code is available at \url{https://github.com/Javion11/PointLiBR.git}.
\end{abstract}

\section{Introduction}
	Point cloud semantic segmentation assigns semantic labels to points within 3D data to enhance the perception, and has diverse applications in many fields like autonomous driving, robotics, and augmented reality. The development trends in point cloud semantic segmentation have focused on improving the robustness and accuracy of segmentation algorithms through the utilization of deep learning architectures\cite{qi2017pointnet, qi2017pointnet++, li2018pointcnn, wu2019pointconv, zhao2021point, han2024largescale}. The remarkable endeavors dedicated to improving network architectures have led to substantial advancements in the performance of point cloud segmentation. Some other works aim to address novel challenges with specific training and testing paradigms, such as few-shot\cite{zhao2021few} and weakly-supervised\cite{xu2020weakly, zhang2021weakly}.
	
	However, these discriminative networks have not taken the detrimental impact of class imbalance into consideration. The categories with more training samples tend to get better segmentation accuracy. Conversely, if there are not sufficient training samples of certain categories, the high-dimensional abstract features of these categories will be obscured within the majority categories. Although oversampling\cite{chawla2002smote} minority categories or undersampling\cite{liu2008exploratory} majority categories can balance the training samples of different categories, processing data with such a brute-force way in point cloud segmentation will compromise scene integrity and disrupt the distribution of data. Another common method for class imbalance is to balance the contributions of samples to the backpropagation gradient by adjusting the loss function, such as Focal Loss\cite{lin2017focal} and Weighted Cross-Entropy(CE) Loss\cite{aurelio2019learning}. Unfortunately, these methods exhibit limited improvements when applied to the task of point cloud semantic segmentation.

\begin{figure}[ht]
\centering
\includegraphics[width=0.99\textwidth]{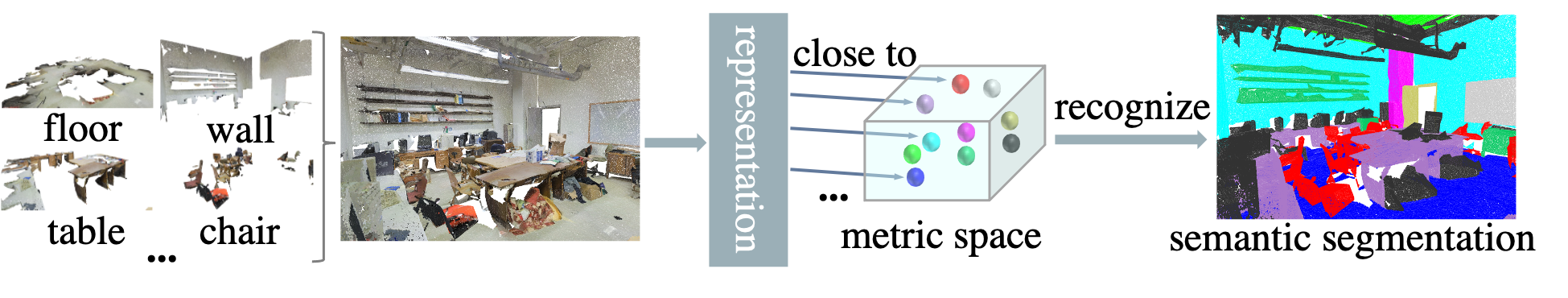} 
\caption{Semantic understanding of \textbf{SPG}. The input point cloud is abstracted into high-dimensional features through a representation network. Then these features are assigned the corresponding category labels, which belong to the closest prototypes in the metric space.}
\label{fig1}
\end{figure}

	To abstract away from the concept of sample quantities, points of the same category are regarded to share inherent uniformity in the subspace prototype guidance (\textbf{SPG}) method, specifically in the sense that their descriptions in the high-dimensional feature space are akin\cite{snell2017prototypical} as shown in \cref{fig1}. When some points are far away from their category prototypes after passing through the representation network, these points are often misclassified into incorrect categories. Therefore, \textbf{SPG} utilizes instance-balanced sampling to ensure scene integrity and facilitate the learning of improved representation networks\cite{kang2019decoupling}. It also employs category prototypes to narrow the intra-class variance in the feature space of the segmentation network, which helps separate the features of minority category from those of majority category. Consequently, it enhances the discrimination of minority category features and facilitates the correct classification by the classifier. 
	
	Since the point sets of different categories coexist in the scene, the features belonging to the categories with few samples tend to be overshadowed in the features belonging to the categories with many samples, which makes it challenging to obtain accurate feature prototypes. \textbf{SPG} utilizes an auxiliary branch composed of an encoder and a projection head to separately process the features of the specific category, which is beneficial to extract the prototypes of minority categories. Furthermore, to prevent the overlap or interference of inter-class features, the input point clouds of the auxiliary branch are preprocessed by grouping them based on their categories. It can separate the mixed feature space, which effectively extracts homogeneous features and mitigates the influence of heterogeneous contextual information. 
	
	The prototype for each category contains information exclusively from points of the same category after the grouping process. \textbf{SPG} widens the distance between heterogeneous features through the loss constraint\cite{khosla2020supervised} to increase the discrimination of the corresponding prototypes. Prototypes containing historical scene information exhibit increased robustness to scene variations. Therefore, prototypes are updated with momentum\cite{he2020momentum,chen2020improved} to retain their ability to recognize different scenes, rather than relying solely on the information of current scene. 
	
	In addition to employing prototypes from the auxiliary branch to guide the training of the main segmentation network, the feature space of the auxiliary branch is also supervised by the correctly classified features from the main branch. The two branches mutually constrain each other and undergo joint online training, which ensures alignment in the convergence direction of the overall network. The main contributions of this work can be summarized as follows:
\begin{itemize}
\item This work analyzes the distribution of the features in the point cloud semantic segmentation network and introduces to use point cloud category prototypes as convergence targets to constrain the feature representation of the segmentation network.
\item A novel method for extracting point cloud prototypes is proposed, which uses a plug-and-play branch consisting of an encoder and a projection head. This branch projects class-grouped point clouds into separate feature subspaces and extracts prototypes that are not interfered by other categories.
\item The proposed method \textbf{SPG} utilizes prototypes from separate subspaces to guide the training of segmentation network. It significantly mitigates the negative impact of class imbalance on point cloud semantic segmentation without increasing the computational overhead during inference.
\end{itemize}

\section{Related Work}
	This paper mitigates class imbalance in point cloud semantic segmentation with the point cloud prototype extracted from separate subspaces. We briefly review recent works on point cloud semantic segmentation, class imbalance, and category prototype.
	
\paragraph{Point cloud semantic segmentation.} Compared to traditional non-learning-based approaches, deep learning-based methods\cite{qi2017pointnet, qi2017pointnet++, li2018pointcnn, zhao2021point, wu2019pointconv, thomas2019kpconv, wu2022point} have elevated the performance of semantic segmentation to a new level. In recent years, newly proposed algorithms\cite{feng2023clustering, saltori2023walking, takmaz20233d, liu2023mars3d, li2023mseg3d, xiao20233d, han2024largescale} for point cloud semantic segmentation have predominantly leveraged deep learning techniques. In these noteworthy works, some focus on data augmentation strategies during training for input point clouds, such as PointNeXT\cite{qian2022pointnext} and Mix3D\cite{Nekrasov213DV}. Meanwhile,  PointNet++\cite{qi2017pointnet++}, PointCNN\cite{li2018pointcnn}, PointConv\cite{wu2019pointconv}, PTv1\cite{zhao2021point}, and PTv2\cite{wu2022point} achieve performance gains by improving the network structure. PointNet\cite{qi2017pointnet} initially introduced a deep learning architecture for handling unordered point cloud data. PointNet++\cite{qi2017pointnet++} introduced hierarchical network architecture; PointCNN\cite{li2018pointcnn} and PointConv\cite{wu2019pointconv} introduced convolutional operations; PTv1\cite{zhao2021point} introduced self-attention mechanisms. Additionally, there are studies that lean towards specific application scenarios\cite{takmaz20233d, li2023mseg3d, xiao20233d, yan20222dpass} or different training paradigms\cite{zhao2021few, xu2020weakly, zhang2021weakly}.

\paragraph{Class imbalance.} The class imbalance problem refers to the situation where there is a significant disparity in the number of samples among different categories. This imbalance will adversely affect the performance of neural networks\cite{zhang2023deep}. To address the class imbalance problem in deep learning, various techniques and strategies have been proposed, including resampling\cite{he2009learning, chawla2002smote, liu2008exploratory, ren2020balanced, zhou2020bbn}, weighted loss functions\cite{lin2017focal, dong2018imbalanced, aurelio2019learning, shu2019meta}, generative methods\cite{ren2019ewgan, wang2020deep}, transfer learning\cite{yin2019feature, liu2020deep}, and re-calibration\cite{tian2020posterior, pan2021model}. Due to the complexity and sparsity of point clouds, the above mentioned methods can hardly address class imbalance issues in point cloud semantic segmentation. 

\paragraph{Category prototype.} In few-shot learning\cite{vinyals2016matching, snell2017prototypical, ren2018meta, sung2018learning, li2019few}, prototypes are often utilized as representatives of categories with limited training samples, aiding in the generalization to new instances or unseen categories with few examples. These prototypes can be used to measure the similarity between query instances and support instances belonging to different categories, enabling effective classification or regression. In weakly-supervised learning\cite{li2020prototypical, xu2020weakly, zhang2021weakly, chen2022self, du2022weakly}, prototypes can be employed to represent latent structures or concepts within the data, guiding the learning process in the absence of precise labels or annotations. In both few-shot and semi-supervised learning, the problems are addressed within fixed training and testing paradigms. However, minority category prototypes still tend to be influenced by prototypes of other categories, thereby leaving the issue of class imbalance unresolved. 

\section{Motivation} 

\begin{figure}[ht]
\centering
\includegraphics[width=0.80\textwidth]{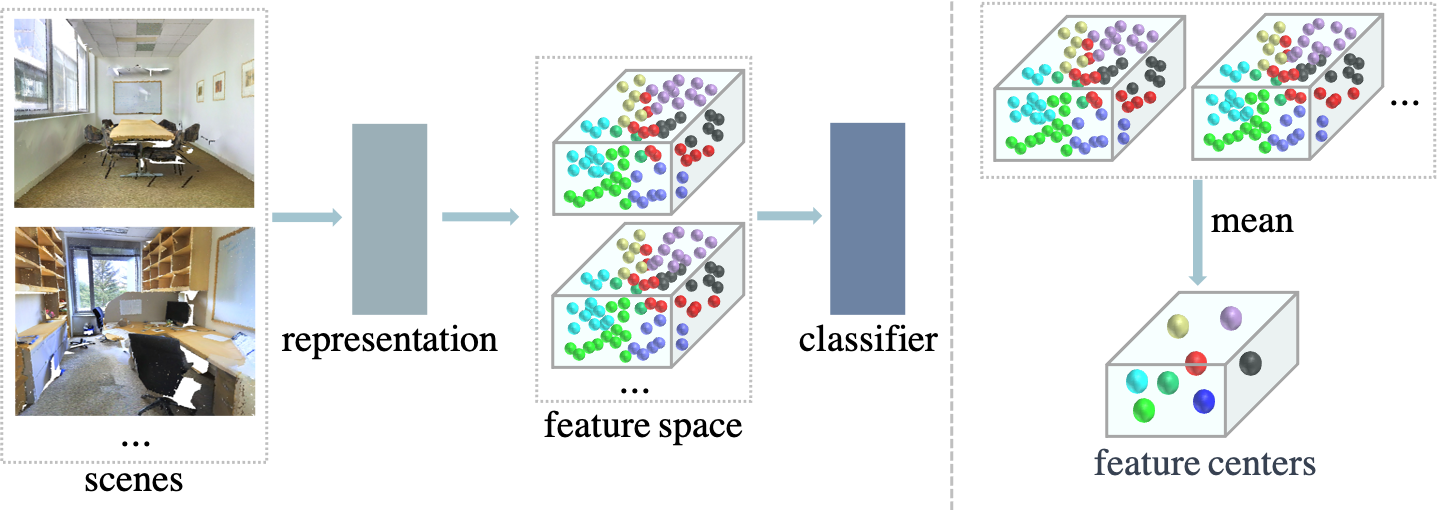} 
\caption{The general process of point cloud semantic segmentation networks (including PointNet++\cite{qi2017pointnet++}, PTv1\cite{zhao2021point}, PTv2\cite{wu2022point}, RandLANet\cite{hu2020randla}, KPConv\cite{thomas2019kpconv}, etc.). }
\label{fig2}
\end{figure}

\begin{table}[ht]\scriptsize
\setlength\tabcolsep{2pt}
\centering
 \caption{The cosine similarity between the feature centers of TP (True Positive), FP (False Positive), and FN (False Negative) subsets in the test set and the feature center of the TP subset in the training set.} 
 \label{tab}
 \begin{tabular}{c|ccccccccccccc}
\toprule 
 Category  & ceiling    & floor      & wall      & column     & window & door       & table & chair & sofa       & bookcase & board & clutter \\ \midrule
 \rowcolor[gray]{0.8} TP & 1.00  &1.00 & 1.00  &1.00 & 1.00  &1.00 & 1.00  &1.00 & 1.00  &1.00 & 1.00  &1.00   \\ 
 FP & 0.94  &0.94 & 0.98  &0.98 & 0.97  &0.97 & 0.98  &0.97 & 0.93 &0.99 & 0.97  &0.99   \\ 
 FN & 0.55 & 0.64 & 0.78 & 0.75 & 0.74 & 0.78 & 0.78 & 0.75 & 0.81 & 0.82 & 0.85 & 0.81 \\ \bottomrule
 \end{tabular}
\end{table}

\begin{figure}[ht]
\centering
\includegraphics[width=0.82\textwidth]{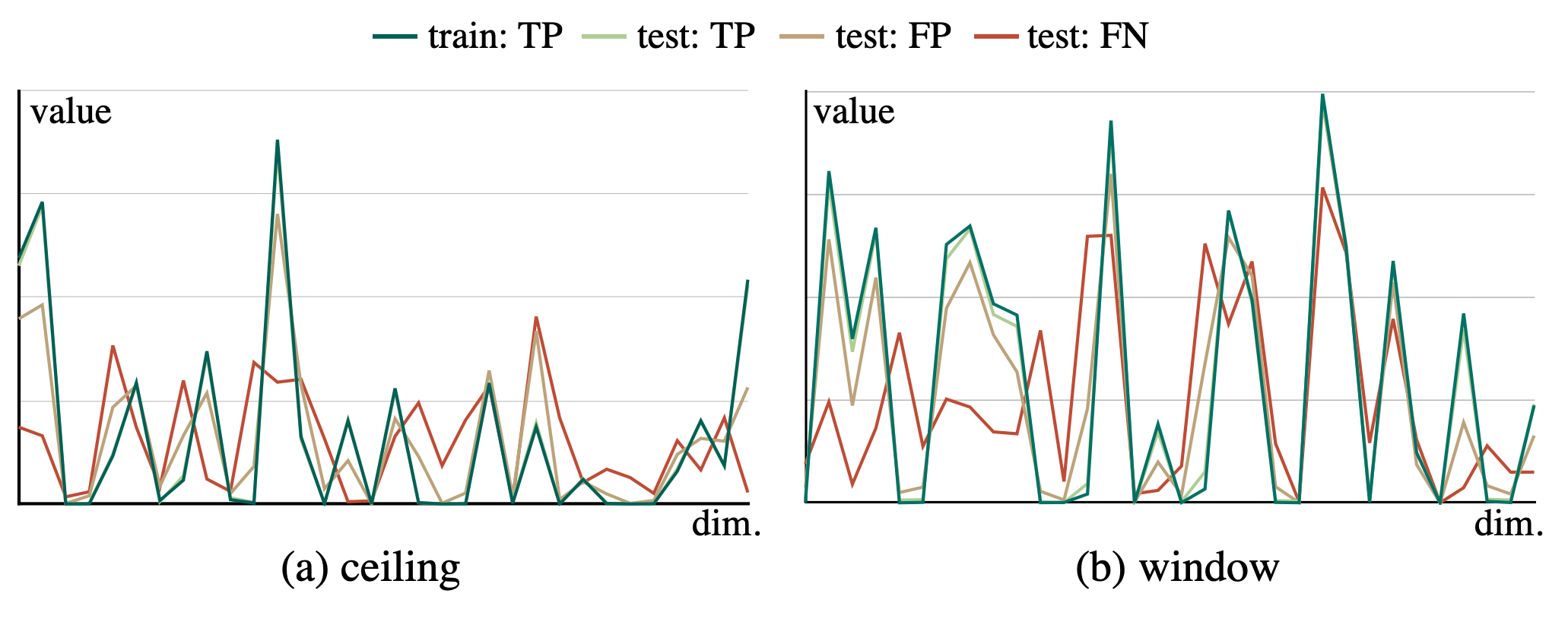} 
\caption{The values of feature centers of "ceiling" and "window" in each dimension.}
\label{fig3}
\end{figure}

\begin{figure}[ht]
\centering
\includegraphics[width=0.96\textwidth]{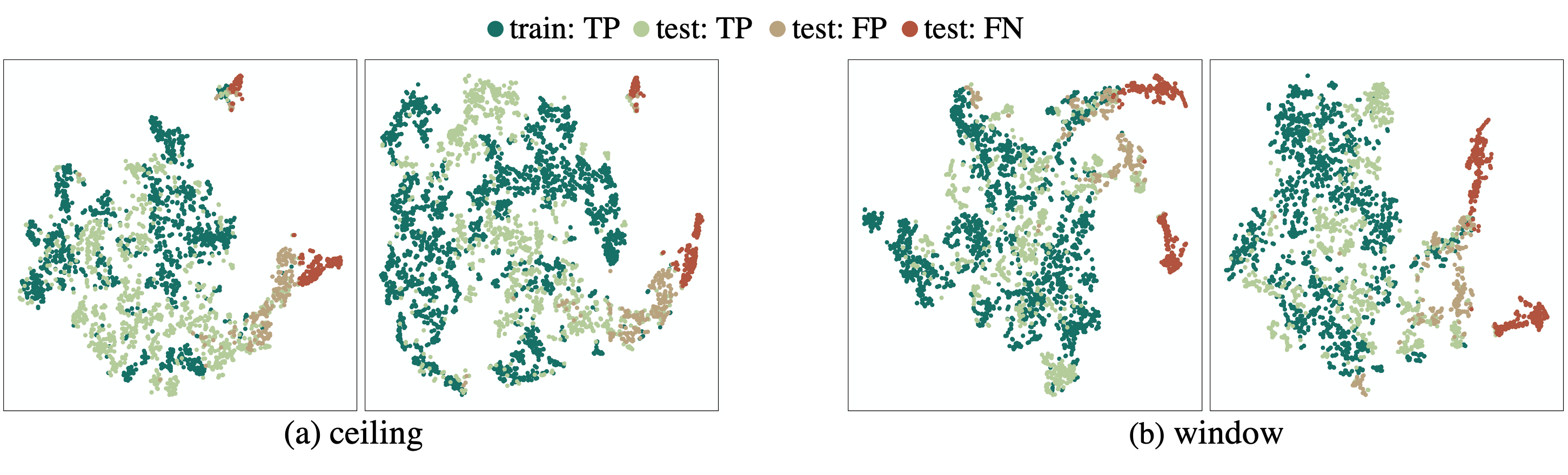} 
\caption{The feature distributions of "ceiling" and "window". They are visualized by t-SNE\cite{van2008visualizing}.}
\label{fig4}
\end{figure}

\begin{figure*}[ht]
\centering
\includegraphics[width=0.98\textwidth]{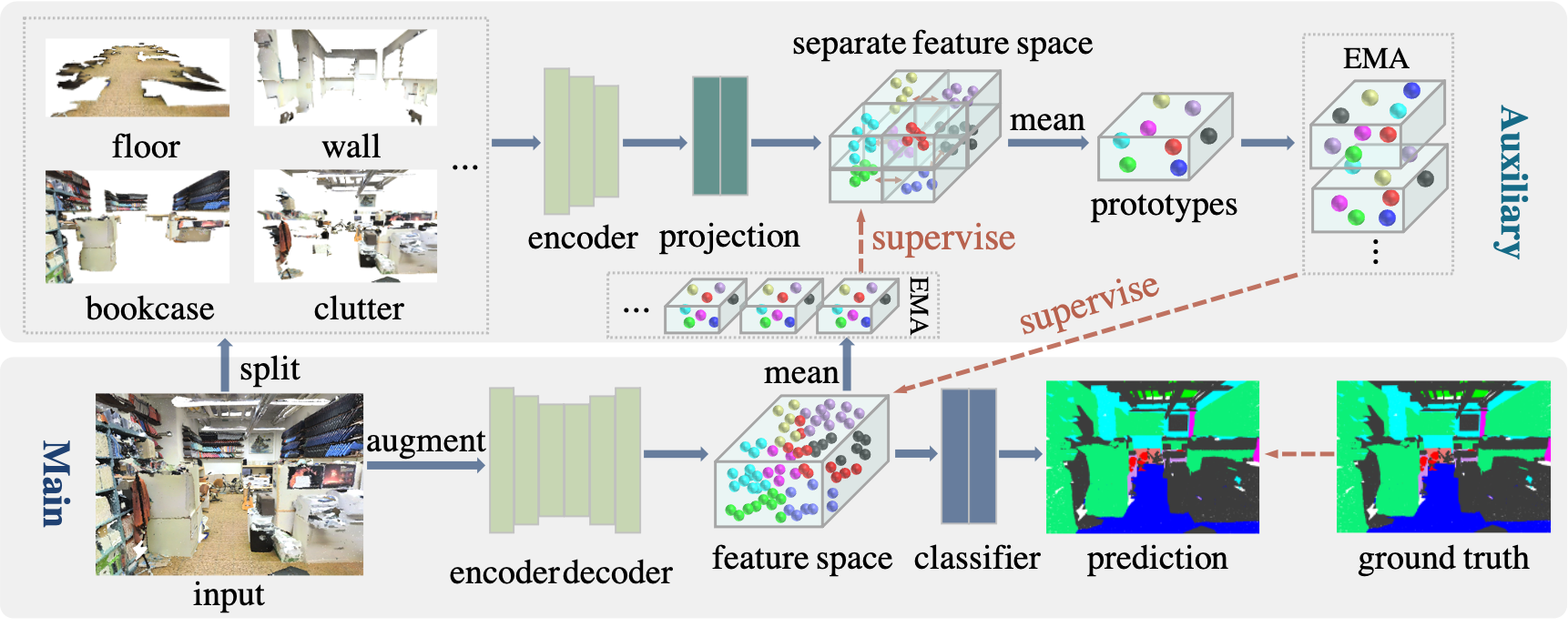} 
\caption{The overall framework of \textbf{SPG}. The grouped point sets are projected into separate feature subspaces via a plug-and-play auxiliary branch, facilitating the derivation of discriminative prototypes. The feature prototypes of the current scene, in combination with those from historical scenes, provide supervision for the feature space of the main branch to reduce the intra-class variance. Simultaneously, the features correctly classified by the main branch also supervise the feature space of the auxiliary branch, ensuring their convergence aligning with each other.}
\label{fig5}
\end{figure*}

	A typical point cloud semantic segmentation network consists of a representation and a classifier, as shown in \cref{fig2}. The representation maps the input point cloud to a high-dimensional feature space, and then the features are classified by the classifier. \cref{tab} and \cref{fig3} illustrate illustrate the relationships between feature centers of different point subsets in the feature space of the PTv1, which is trained and tested on the S3DIS dataset\cite{armeni20163d}. The feature centers of TP points in the training set and test set are closely aligned, while the feature centers of FN points in the test set are obviously different from those of TP points. In \cref{fig4}, when some point features are far away from the feature centers, they may be misclassified. Furthermore, we validated through experiments that PointNet++, PTv2, RandLANet, and KPConv exhibit similar phenomenon in their feature spaces. 
	
	Therefore, the reason why point cloud segmentation networks can correctly understand scenes is that they project the point cloud with complex textures and spatial information into the vicinity of corresponding feature prototypes in the metric space as shown in \cref{fig1}. The discriminative prototypes for minority categories and small intra-class variance will lead to better performance on minority categories. So how to effectively obtain and utilize high-quality prototypes for minority categories becomes the key issues in this study.

\section{Methodology}
	The overview of the proposed method \textbf{SPG} is shown in \cref{fig5}. \textbf{SPG} consists of a dual-branch structure, with the main branch being the mainstream segmentation network such as PointNet++, PTv1, PTv2, and the auxiliary branch comprising the encoder from the segmentation network along with a projection head. The details of \textbf{SPG} will be introduced in this section, including how to extract feature prototypes of point clouds by the auxiliary branch, how to utilize the prototypes to supervise the main segmentation network and how to align the convergence direction of the two branches. Suppose that $\mathbf{X}$ is the point set of current scene. The task of \textbf{SPG} is to assign a per point label to each member of $\mathbf{X}$, which will be realized by the main branch. Hence, \textbf{SPG} only requires the main branch to infer the point cloud, and it does not introduce any additional computational complexity compared to the base network. \textbf{SPG} allows various choices of the base network without any constraints.

\subsection{Point Cloud Prototype Extraction}
	The category prototype extraction places a stronger emphasis on the intrinsic characteristics of different point categories. To avoid introducing interference from heterogeneous features when the encoder aggregates neighborhood information, the point cloud $\mathbf{X}$ of whole scene is separated into several point sets $\{ \mathbf{X}_1, \cdots, \mathbf{X}_c, \cdots, \mathbf{X}_C\}$ by category labels, where the subscript $c$ refers to category $c$ and the capital $C$ represents the number of categories. Subsequently, these point sets are input into the auxiliary branch network, which is composed of the encoder $f_e(\cdot)$ from the base segmentation network and a projection head $f_p(\cdot)$. The projection head is a multi-layer perceptron (MLP) with one hidden layer. 
	
	The high-dimensional abstract feature set $\mathbf{H}_c$ of category $c$ are extracted by the encoder $f_e(\cdot)$, which could be formalized as:
\begin{equation}
\mathbf{H}_c = f_e(\mathbf{X}_c).
\label{equ1}
\end{equation}
Then, the projection head is used to map the high-dimensional abstract feature $\mathbf{h}_{c,i} \in \mathbf{H}_c$ to the space $\mathcal{S}_c$ where the corresponding category prototype is extracted, where $i$ means the feature index. The projection head is a multi-layer perceptron (MLP) with one or two hidden layers depending on the feature dimension. The MLP with one hidden layer can be expressed by the following equation:
\begin{equation}
\mathbf{f}_{c,i} = L_2(\sigma^{(2)}(W^{(2)} \sigma^{(1)} (W^{(1)}\mathbf{h}_{c,i}))), 
\label{equ2}
\end{equation}
where $L_2(\cdot)$ denotes $L_2$ normalization; $\sigma^{(1)}$ and $\sigma^{(2)}$ are the ReLU nonlinearities; $W^{(1)}$ and $W^{(2)}$ are the weight of linear layers.  A supervised contrastive loss $L_{con}$ between point features is employed to enhance the discrimination of features from different categories. 
\begin{equation}
L_{con} = \sum_{c=1}^{C} \frac{-1}{N_c-1} \sum_{i=1}^{N_c }\sum_{p\in P(i)} log\frac{exp(\mathbf{f}_{c,i}\cdot \mathbf{f}_{c,p}/\tau )}{\sum_{a\in A(i)}exp(\mathbf{f}_{c,i}\cdot \mathbf{f}_a/\tau )},
\label{equ3}
\end{equation}
where $N$ and $N_c$ refer to the number of features in separate space $\mathcal{S}_c$ and entire space $\mathcal{S}$, respectively; $\tau$ is the scalar temperature parameter whose default value is $0.07$; $A(i)\equiv \{1,2,\cdots, N\} \setminus \{i\}$, and $P(i)\equiv \{1,2,\cdots, N_c\} \setminus \{i\}$. It furnishes a portion of the backpropagation gradient for updating the auxiliary branch network, and the other portion is supplied by prototype supervision from the feature space of main branch, which is depicted in \cref{fig5}.

\noindent
\begin{minipage}{\textwidth}
        	\begin{minipage}[h]{0.55\textwidth}
        \centering
        \includegraphics[height=0.7\textwidth]{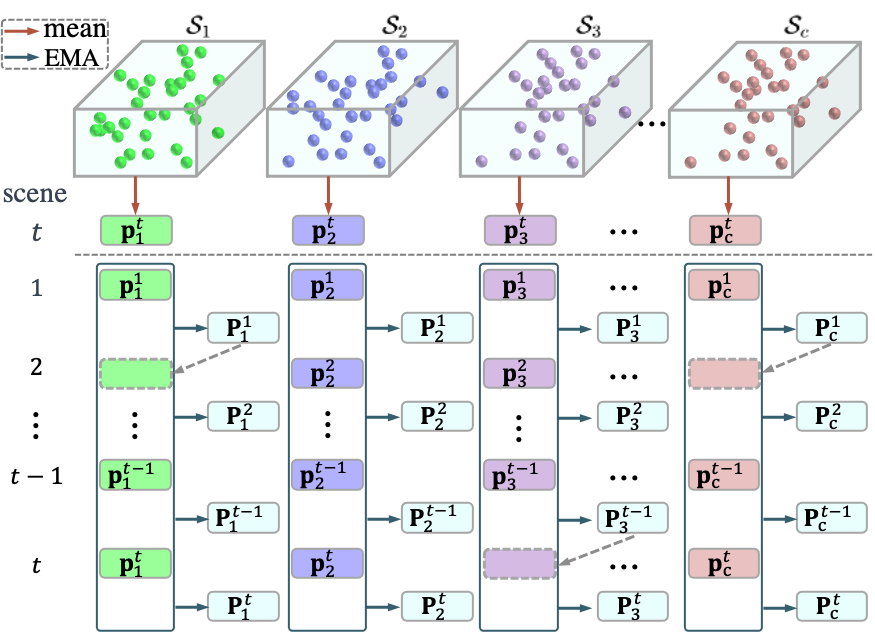}
        \makeatletter\def\@captype{figure}\makeatother\caption{The extraction of prototypes from separate feature subspaces and the EMA process. If there is no point set for category $c$ in the current scene $t$, its prototype is assigned with the EMA of the prototypes from previous $t-1$ scenes.}
        \label{fig6}  
        \end{minipage}
        \begin{minipage}[h]{0.44\textwidth}
        	\centering
        \includegraphics[height=0.5\textwidth]{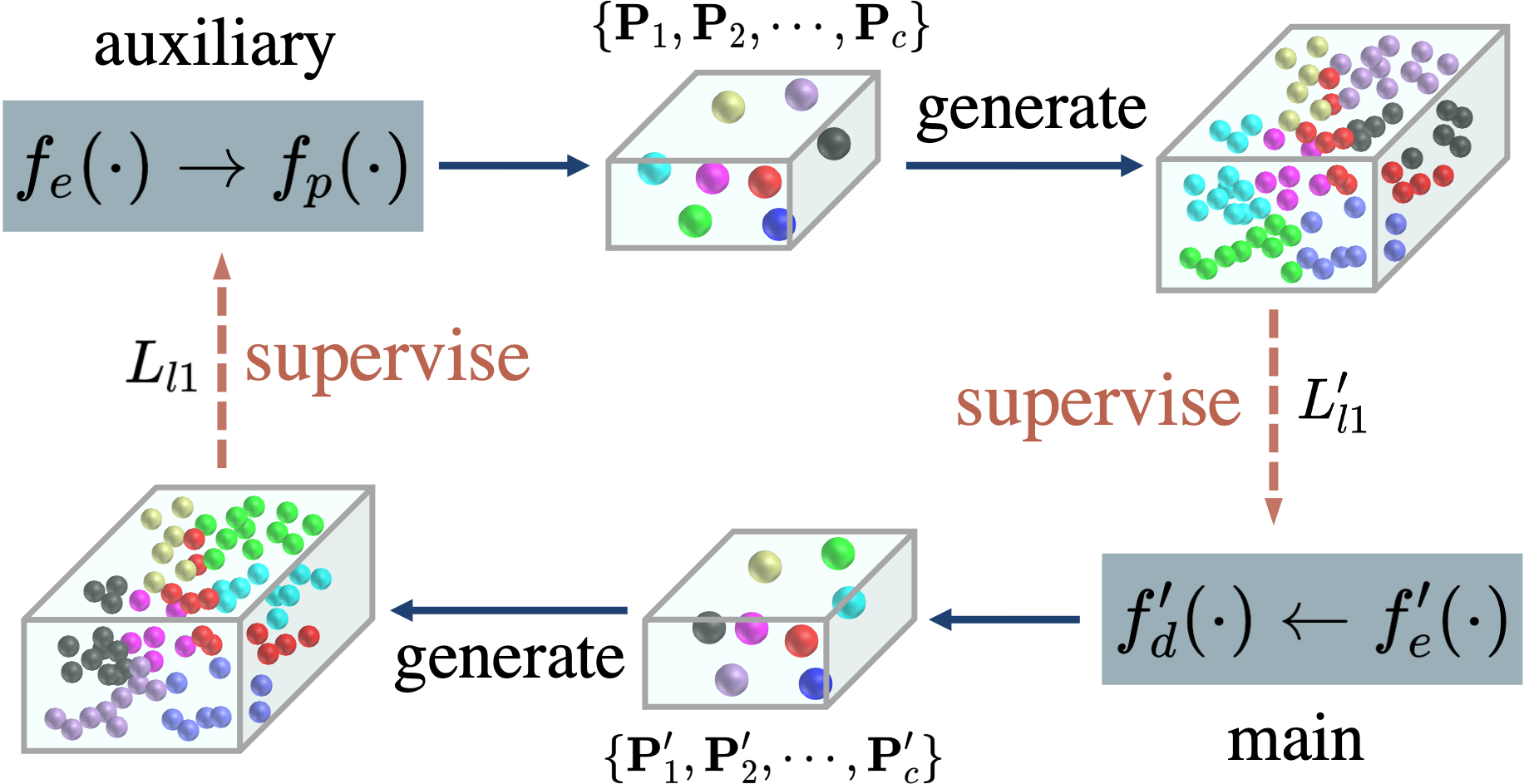}
        \makeatletter\def\@captype{figure}\makeatother\caption{Interaction loop between the auxiliary branch and the main branch. The two branches use their respective prototypes to generate features for mutual supervision, transferring cognitive prototype knowledge.}
        \label{fig7}
        \end{minipage}
\end{minipage}

	\cref{fig6} illustrates the extracting process of prototypes $\{\mathbf{P}_1^t, \mathbf{P}_2^t, \cdots, \mathbf{P}_c^t\}$ from separate feature subspaces $\{\mathcal{S}_1, \mathcal{S}_2, \cdots, \mathcal{S}_c\}$ in the auxiliary. The prototype $\mathbf{p}_c^t$ of the current scene $t$ for category $c$ in the feature subspace $\mathcal{S}_c$ is obtained by taking the mean of normalized features of category $c$, which could be formalized as:
\begin{equation}
\mathbf{p}_c^t = \frac{1}{N_c}\sum_{i=1}^{N_c}\mathbf{f}_{c,i}^t.
\label{equ4}
\end{equation}
In order to ensure that the prototypes extracted by the auxiliary branch contain not only the semantic information of current scene but also the collective semantic information from all training scenes, exponential moving average (EMA) is applied to obtain the prototypes:
\begin{equation}
\mathbf{P}_c^t = (1-\alpha)\mathbf{p}_c^t + \alpha\mathbf{P}_c^{t-1},
\label{equ5}
\end{equation}
where the parameter $\alpha$ represents the smoothing factor, controlling the weight of the new observation $\mathbf{p}_c^t$ in calculating $\mathbf{P}_c^t$. 
	
\subsection{Consistency Constraint for Convergence}
	The input $X$ is processed through the encoder $f_e^\prime(\cdot)$ and decoder $f_d^\prime(\cdot)$ of the main branch to obtain the feature set $\mathbf{H}^\prime$:
\begin{equation}
\mathbf{H}^\prime = f_d^\prime(f_e^\prime(\mathbf{X})).
\label{equ6}
\end{equation}
	The discriminative prototypes $\{\mathbf{P}_1, \mathbf{P}_2, \cdots, \mathbf{P}_c\}$, which is obtained from the separate feature subspaces in the auxiliary branch, are used to guide the distribution of $\mathbf{H}^\prime$. This process can be formalized as follows:
\begin{equation}
L_{l1}^{\prime} = \frac{1}{N^\prime}\sum_{c=1}^{C}\sum_{i\in \mathcal{P}(c)} \mathrm{smooth}_{L1}(L_2(\mathbf{h}_i^{\prime}) - \mathbf{P}_{c}), 
\label{equ7}
\end{equation}
in which
\begin{equation}
\operatorname{smooth}_{L_{1}}(x)=\left\{\begin{array}{ll}
0.5 x^{2} & \text { if }|x|<1, \\
|x|-0.5 & \text { otherwise }.
\end{array}\right.
\end{equation}
Here, $N^\prime$ is the number of features in $\mathbf{H}^\prime$; $\mathcal{P}(c)$ refers to the index set of the feature subset that belongs to category $c$; $\mathbf{h}_i^{\prime}$ is the $i$-th feature in $\mathbf{H}^\prime$.
	
	The loss $L_{l1}^{\prime}$ in \cref{equ7} and cross-entropy loss $L_{ce}^{\prime}$ jointly provides the backpropagation gradient for the update of the segmentation network in main branch. The cross-entropy loss is expressed as follows:
\begin{equation}
L_{ce}^{\prime} = -\frac{1}{N^\prime}\sum_{i=1}^{N^\prime}y_{i}log(f_{c}(\mathbf{h}_i^{\prime})),
\label{equ9}
\end{equation}
where $f_{c}(\cdot)$ represents the classifier and $y_{i}$ is the ground truth label for the $i$-th point. 

	As shown in \cref{fig7}, \textbf{SPG} not only employs the feature prototypes of auxiliary branch to constrain the feature space of main branch (\cref{equ7}), but also utilizes the feature prototypes of main branch to constrain the feature space of auxiliary branch for the consistent convergence between the main branch and the auxiliary branch. The loss $L_{l1}$ in \cref{fig7} could be expressed as:
\begin{equation}
L_{l1} = \frac{1}{N}\sum_{c=1}^{C}\sum_{i\in \mathcal{P}^\prime(c)} \mathrm{smooth}_{L1}(\mathbf{f}_{c,i} - \mathbf{P}_{c}^{\prime}), 
\label{equ10}
\end{equation}
where $\mathcal{P}^{\prime}(c)$ refers to the index set of the features in $\mathcal{S}_c$. The prototypes of main branch, denoted as $\{\mathbf{P}_1^\prime, \mathbf{P}_2^\prime, \cdots, \mathbf{P}_c^\prime\}$, are extracted from the feature sets $\{\mathbf{H}^{\prime 1}, \mathbf{H}^{\prime 2}, \cdots, \mathbf{H}^{\prime t}\}$ in the same manner as described in \cref{equ4} and \cref{equ5}. To ensure feature accuracy, only features that are classified correctly are selected from $\mathbf{H}^{\prime}$ to participate in prototype computation. 

	Since the two branches of the \textbf{SPG} framework are trained simultaneously online, the overall loss is simply the sum of the losses from \cref{equ3}, \cref{equ10}, \cref{equ7} and \cref{equ9}. This does not introduce complex training strategies, and it is represented as follows:
\begin{equation}
L = w_1L_{con} + w_2L_{l1} + w_3L_{l1}^\prime + w_4L_{ce}^\prime, 
\label{equ11}
\end{equation}
where $w_1$, $w_2$, $w_3$ and $w_4$ are the loss weights and their default values are $1$. 

%
	
\section{Experiment}
	The experimental evaluations are conducted on four public benchmark datasets, S3DIS\cite{armeni20163d}, ScanNet v2\cite{dai2017scannet}, ScanNet200\cite{rozenberszki2022language} and Toronto-3D\cite{tan2020toronto}, and the collected real-world data. The effectiveness of the proposed method are validated through relevant experiments. The real-world data is outdoor point cloud with color information, collected by \textbf{Livox Horizon LiDAR} and fused with RGB images. 
\subsection{Data and Metric}
	All of these four publicly available datasets have been widely used to evaluate the performance of 3D semantic segmentation algorithms. S3DIS consists of point clouds captured from 6 areas with detailed annotations for 13 semantic categories. ScanNet v2 provides 1513 RGB-D scans of various scenes with per-point semantic labels for 20 categories, in which 1201 scenes are used for training and 312 for validation. ScanNet200 consists of 200 categories, an order of magnitude more than previous. Toronto-3D covers 1km of streets and consists of about 78.3 million points, which is divided into four sections. The second section is used for validation and the rest sections for training. For evaluating the performance of the proposed method on S3DIS area5, the metrics of mean class-wise intersection over union (mIoU), mean of class-wise accuracy (mAcc), and overall point-wise accuracy (OA) are utilized. Following a standard protocol\cite{qi2017pointnet++, wu2022point}, mIoU is employed as the evaluation metric for assessing the performance on the validation set of ScanNet v2 and ScanNet200. Toronto-3D adopts the metrics of mIoU and OA.

\begin{table}[ht] \tiny
\centering
\caption{Semantic segmentation performance on S3DIS Area5, ScanNet v2, ScanNet200 and Toronto-3D.}
\label{tab1}
\begin{tabular}{c|cc|ccc|c|c|cc}
\toprule
\multirow{2}{*}{Method} & \multirow{2}{*}{input} & \multirow{2}{*}{Params.(M)} & \multicolumn{3}{c|}{S3DIS Area5} & ScanNet v2 & ScanNet200 & \multicolumn{2}{c}{Toronto-3D}\\
           &       &      & OA & mACC & mIoU & mIoU & mIoU &OA & mIoU\\ \midrule
PointCNN\cite{li2018pointcnn}   & point & 0.6  & 85.9   & 63.9     & 57.3     & -    & -   & - & - \\
DGCNN\cite{wang2019dynamic} & point &1.3 & 83.6 & - & 57.9 & -  &- & 94.2 & 61.8 \\
PointConv\cite{wu2019pointconv}  & point & -    & -      & 67.0     & 58.3     & 61.0  & - & - & -  \\
KPConv\cite{thomas2019kpconv}    & point & 15.0 & -      & 72.8     & 67.1     & 69.2  & -   & 95.4 & 69.1 \\
SparseConvNet\cite{graham20183d} & voxel & -  & -      & -     & -     & 69.3   & - & - & -  \\
MinkowskiNet\cite{choy20194d}   & voxel & -    & -      & 71.7     & 65.4     & 72.1  & 25.0 & - & -  \\
CBL\cite{tang2022contrastive} & point & 18.6 & 90.6      & 75.2     & 69.4     & -    & - & - & - \\
PointNeXt\cite{qian2022pointnext}  & point      & 41.6    & 90.6    & 76.8      & 70.5      & -    & - & -   \\     
\makecell{Stratiﬁed \\Transformer\cite{lai2022stratified}} & point & 8.0 & 91.5      & 78.1     & 72.0     & 74.3   & -  & - & - \\ \midrule
PointNet++\cite{qi2017pointnet++} & point & 1.0  & 83.0   & -        & 53.5     & 53.5   & -   & 92.6 & 59.5 \\ 
\rowcolor[gray]{0.8} &  &   & 87.6   &        & 64.0     & 55.8   &    & 93.8 & 61.2  \\
\rowcolor[gray]{0.8} \multirow{-2}{*}{\textbf{SPG}(PointNet++)} & \multirow{-2}{*}{point} & \multirow{-2}{*}{1.0} & ($\uparrow$\textbf{4.6}) &  \multirow{-2}{*}{70.8} & ($\uparrow$\textbf{11.5}) & ($\uparrow$\textbf{2.3}) &\multirow{-2}{*}{-} & ($\uparrow$\textbf{1.2}) & ($\uparrow$\textbf{1.7})  \\  \midrule
 
PTv1\cite{zhao2021point}       & point & 7.8  & 90.8   & 76.5     & 70.4     & 70.6   & 27.8   & 96.1  & 78.8 \\
\rowcolor[gray]{0.8} &  &   & 91.2   & 77.9       & 71.5     & 71.3   &  29.1  & 96.7 & 80.0  \\
\rowcolor[gray]{0.8} \multirow{-2}{*}{\textbf{SPG}(PTv1)} & \multirow{-2}{*}{point} & \multirow{-2}{*}{7.8} & ($\uparrow$\textbf{0.4}) &  ($\uparrow$\textbf{1.4}) & ($\uparrow$\textbf{1.1}) & ($\uparrow$\textbf{0.7}) &($\uparrow$\textbf{1.3}) & ($\uparrow$\textbf{0.6}) & ($\uparrow$\textbf{1.2})  \\  \midrule

PTv2\cite{wu2022point}       & point & 3.9/11.3  & 91.1   & 77.9     & 71.6     & 75.4  & 30.2   & 97.2 & 81.6\\ 
\rowcolor[gray]{0.8} &  &   & 91.9   & 79.5       & 73.3     & 76.0   &  31.5  & 97.6 & 83.1  \\
\rowcolor[gray]{0.8} \multirow{-2}{*}{\textbf{SPG}(PTv2)} & \multirow{-2}{*}{point} & \multirow{-2}{*}{3.9/11.3} & ($\uparrow$\textbf{0.8}) &  ($\uparrow$\textbf{1.6}) & ($\uparrow$\textbf{1.7}) & ($\uparrow$\textbf{0.6}) &($\uparrow$\textbf{1.3}) & ($\uparrow$\textbf{0.4}) & ($\uparrow$\textbf{1.5})  \\
\bottomrule
\end{tabular}
\end{table}

\begin{table*}[ht] \tiny
\caption{Semantic segmentation performance of each category on S3DIS Area5.}
\label{tab2}
\centering
\begin{tabular}{c|ccccccccccccc|c}
\toprule
Category      & ceiling    & floor      & wall       & beam & column     & window & door       & table & chair & sofa       & bookcase & board & clutter   & \multirow{3}{*}{mIoU}       \\ \cmidrule{1-14}
Train(\%) & 19.14 & 16.51 & 27.25 & 2.42 & 2.13 & 2.12 & 5.48 & 3.24 & 4.07 & 0.49 & 4.71  & 1.26 & 11.19 &      \\ 
Test(\%)  & 19.57 & 16.54 & 29.20  & 0.03 & 1.76 & 3.52 & 3.03 & 3.75 & 1.87 & 0.27 & 10.35 & 1.19 & 8.92  &      \\ \midrule
PointNet\cite{qi2017pointnet}  & 88.8        & 97.3      & 69.8     & \textbf{0.1}     & 3.9       & 46.3       & 10.8     & 59.0      & 52.6      & 5.9   & 40.3  & 26.4    & 33.2 & 41.1   \\
PointCNN\cite{li2018pointcnn}  & 92.3  & 98.2  & 79.4  & 0.0  & 17.6 & 22.8 & 62.1 & 74.4 & 80.6 & 31.7 & 66.7  & 62.1 & 56.7  & 57.3 \\
KPConv\cite{thomas2019kpconv}    & 92.8  & 97.3  & 82.4  & 0.0  & 23.9 & 58.0 & 69.0 & 81.5 & 91.0 & 75.4 & 75.3  & 66.7 & 58.9  & 67.1 \\
CBL\cite{tang2022contrastive} & 93.9        & 98.4      & 84.2     & 0.0     & 37.0       & 57.7    & 71.9   & 91.7    & 81.8    & 77.8    & 75.6 & 69.1& 62.9   & 69.4     \\
PointNeXt\cite{qian2022pointnext}      & 94.2    & 98.5  & 84.4   & 0.0     & 37.7    & 59.3   & 74.0     & 83.1  & 91.6      & 77.4    & 77.2     & 78.8      & 60.6 & 70.5 \\ \midrule
PTv1\cite{zhao2021point}       & 94.0  & 98.5  & 86.3  & 0.0  & 38.0 & 63.4 & 74.3 & 89.1 & 82.4 & 74.3 & 80.2  & 76.0 & 59.3  & 70.4 \\
\rowcolor[gray]{0.8}  & 95.0 & 98.0 & 87.0 & 0.0 & 44.8 & 63.0  & 73.9 & 83.3  & 91.0 & 77.2 & 77.4     & 76.9  & 61.6 & 71.5 \\ 
\rowcolor[gray]{0.8} \multirow{-2}{*}{\textbf{SPG}(PTv1)}   & ($\uparrow$1.0) & ($\downarrow$0.5) & ($\downarrow$0.7) & (--) & ($\uparrow$6.8) & ($\downarrow$0.4)   & ($\downarrow$0.4) & ($\downarrow$5.8)  & ($\uparrow$8.6) & ($\uparrow$2.9) & ($\downarrow$2.8)    & ($\uparrow$0.9)  & ($\uparrow$1.3) & ($\uparrow$1.1) \\ \midrule
PTv2\cite{wu2022point}      & 95.2  & 98.7  & 86.1 & 0.0 & 38.9 & 61.8 & 72.5 & 83.9 & 92.5 & 75.7 & 77.9  & 83.9 & 63.6  & 71.6 \\ 
\rowcolor[gray]{0.8}  & 95.1 & 98.7 & 86.9 & 0.0 & 38.9 & 64.6  & 78.0 & 85.2  & 92.7 & 83.0 & 79.5     & 85.1  & 65.0 & 73.3 \\ 
\rowcolor[gray]{0.8} \multirow{-2}{*}{\textbf{SPG}(PTv2)}   & ($\downarrow$0.1) & (--) & ($\uparrow$0.8) & (--) & (--) & ($\uparrow$2.8)   & ($\uparrow$5.5) & ($\uparrow$1.3)  & ($\uparrow$0.2) & ($\uparrow$7.3) & ($\uparrow$1.6)    & ($\uparrow$1.2)  & ($\uparrow$1.4) & ($\uparrow$1.7) \\ \bottomrule
\end{tabular}
\end{table*}

\begin{table*}[ht] \tiny
\setlength\tabcolsep{1pt}
\caption{Segmentation performance of each category on Toronto-3D}
\label{tab-toronto}
\centering
\begin{tabular}{c|cccccccc|c}
\toprule
Category      & Road    & Road marking      & Natural      & Building & Utility line     & Pole & Car    & Fence  & \multirow{3}{*}{mIoU}    \\ \cmidrule{1-9}
Train(\%) &42.57	&1.69	&20.63	&24.50	&1.35	&1.57	&6.78	&0.90     &\\ 
Test(\%)  & 63.13	&2.97	&19.24	&8.85	&0.85	&1.54	&3.23	&0.18     & \\ \midrule
PointNet++\cite{qi2017pointnet++}  &92.9	&0.0	&86.1	&82.2	&61.0	&62.8	&76.4	&14.4 & 59.5 \\
DGCNN\cite{wang2019dynamic}    & 93.9	& 0.0	& 91.3	& 80.4	& 62.4	& 62.3	& 88.3	& 15.8  & 61.8 \\
KPConv\cite{thomas2019kpconv}       & 94.6	&0.1	&96.1	&91.5	&87.7	&81.6	&85.7	&15.7 & 69.1  \\
MS-TGNet\cite{tan2020toronto} &94.4	& 17.2	& 95.7	& 88.8	& 76.0	& 74.0	& 94.2	& 23.6  & 70.5  \\ \midrule

PTv1\cite{zhao2021point}                     & 92.9	& 59.6	& 96.7      & 91.9	& 80.6	& 77.4	& 89.1	& 42.2	& 78.8  \\ 
\rowcolor[gray]{0.8} \textbf{SPG}(PTv1) & 94.0($\uparrow$1.1)	& 63.3($\uparrow$3.7)	& 96.7(--)	& 92.4($\uparrow$0.5)	& 82.7($\uparrow$2.1)	 & 79.3($\uparrow$1.9)	& 89.2($\uparrow$0.1)	& 42.4($\uparrow$0.2)	& 80.0 \\ \midrule
PTv2\cite{wu2022point}                          & 96.6	& 72.4	& 96.9	& 91.8	& 86.0	& 81.4	& 90.6	& 37.3	& 81.6  \\ 
\rowcolor[gray]{0.8} \textbf{SPG}(PTv2) & 96.7($\uparrow$0.1)	& 73.8($\uparrow$1.4)	& 97.0($\uparrow$0.1)	& 93.4($\uparrow$1.6)	& 86.4($\uparrow$0.4)	& 81.7($\uparrow$0.3)	& 90.9($\uparrow$0.3)	& 45.1($\uparrow$7.8)	& 83.1 \\ \bottomrule
\end{tabular}
\end{table*}

\begin{figure*}[ht]
\centering
\includegraphics[width=1.0\textwidth]{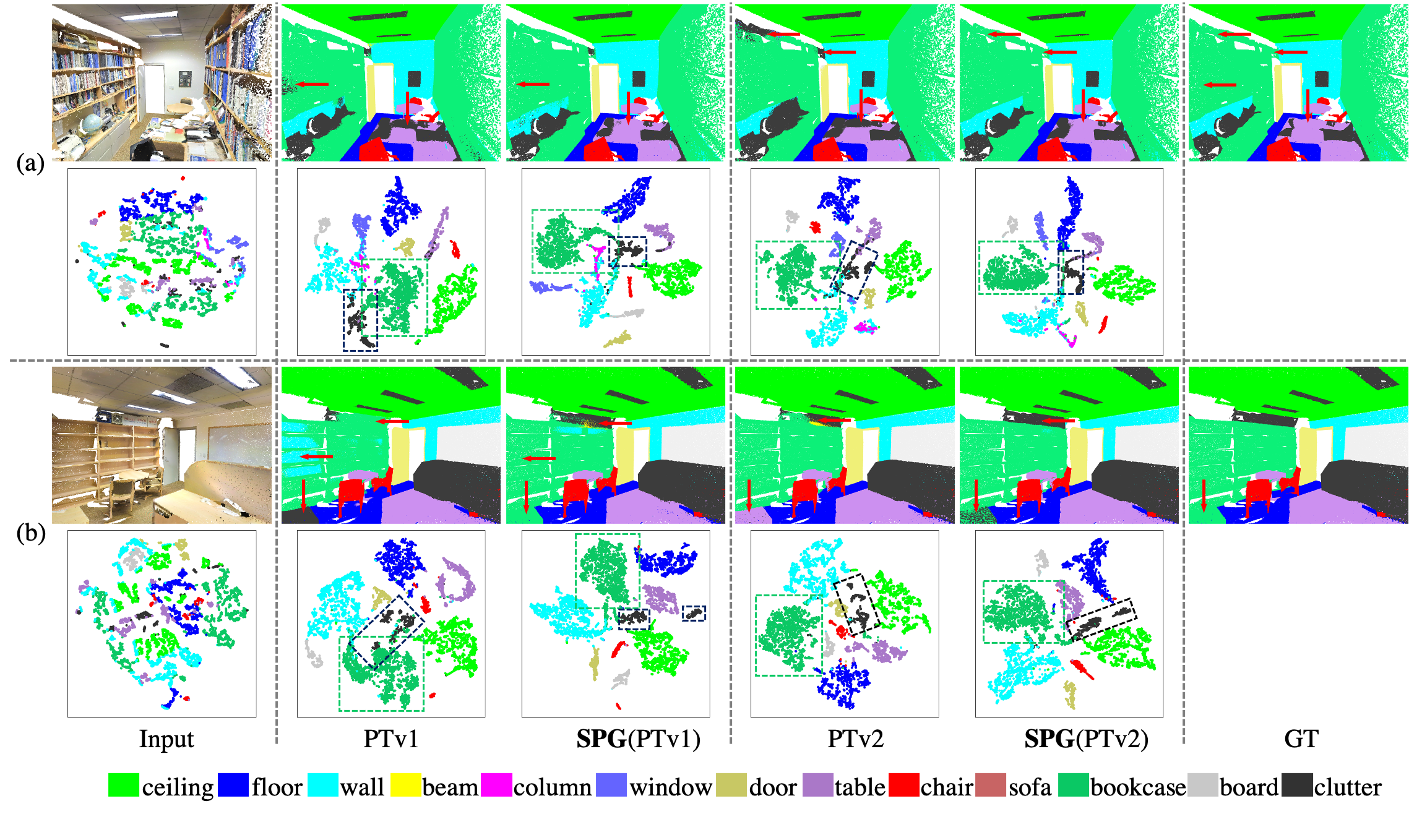}
\caption{Visualization of segmentation results and feature distributions on two scenes in S3DIS Area5. The maps below segmentation results are their corresponding feature distributions. As shown in \cref{fig2}, the first row in (a) and (b) is the visualization of the output segmentation results, and the second row is the visualization of the feature distributions after the representation.} 
\label{fig8}
\end{figure*}

\begin{figure}[ht]
\centering
\includegraphics[width=0.96\textwidth]{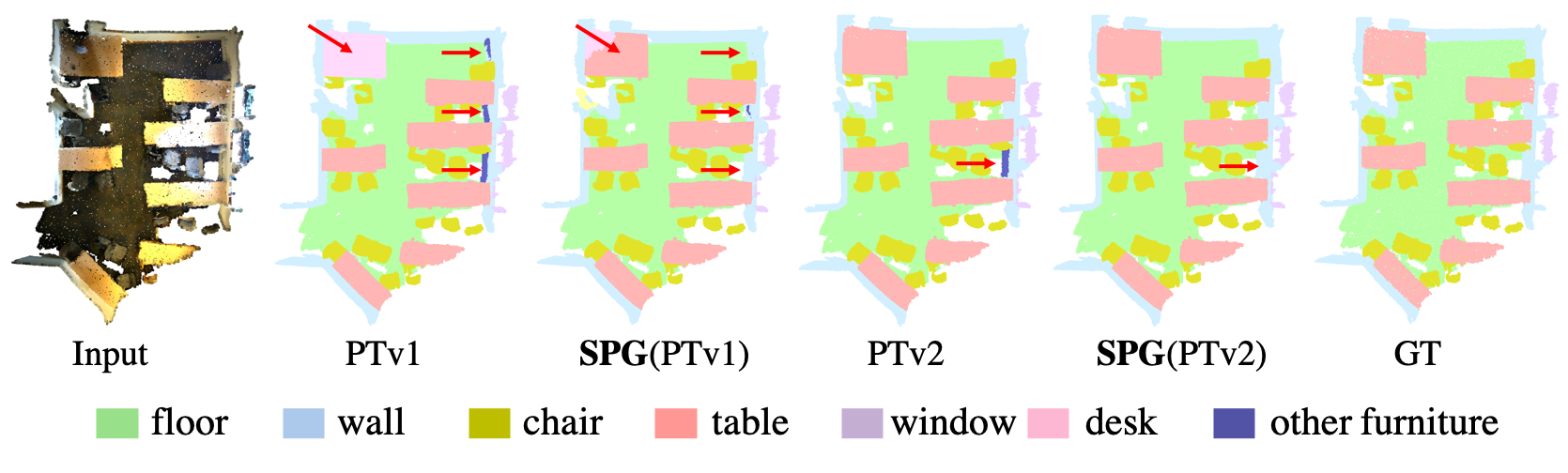}
\caption{Visualization of segmentation results on ScanNet v2. The red arrows point to notable areas.} 
\label{fig9}
\end{figure}

\subsection{Performance Comparison}
	It is evident that \textbf{SPG} significantly improves the performance of base segmentation networks (\ie, PointNet++, PTv1, PTv2) on the four benchmark datasets, as shown in \cref{tab1}. On S3DIS area5, the OA, mAcc, and mIoU metrics for PointNet++ are improved by $4.6\%$ and $11.5\%$; the OA, mAcc, and mIoU metrics for PTv1 are improved by $0.4\%$, $1.4\%$ and $1.1\%$; and the three metrics for PTv2 have been elevated to the state-of-the-art (SOTA) levels, achieving values of $91.9\%$, $79.5\%$, and $73.3\%$, respectively. On the ScanNet v2, the mIoU for the three networks has shown respective improvements of $2.3\%$, $0.7\%$, and $0.6\%$. On the ScanNet200 dataset with a larger number of categories, \textbf{SPG} can also help PTv1 and PTv2 achieve better segmentation performance with a mIoU increase of $1.3\%$. On the outdoor dataset Toronto-3D, \textbf{SPG} can increase the mIoU of PTv1 and PTv2 to more than 80\%, with the latter achieving the top performance on this dataset. The experiments conducted on various datasets with different base networks confirm the strong generalization capability of \textbf{SPG}. Furthermore, this method does not make any modifications to the base segmentation networks during the inference process. So the inference speed of the improved networks will keep the same as the base networks despite the significant increase in performance.
	
	The first two rows in \cref{tab2} and \cref{tab-toronto} refer to the proportions of samples for each category in the training and test datasets, which are measured in terms of the number of points. It can be observed that the segmentation networks exhibit superior performance for the "ceiling", "floor", and "wall" categories on S3DIS and "Road", "Natural", and "Building" categories on Toronto-3D, which have larger sample size. However, the segmentation performance for minority categories is relatively lower. \textbf{SPG} can enhance the segmentation performance of different categories, especially demonstrating more pronounced effects on minority categories. 

\begin{figure}[ht]
\centering
\includegraphics[width=0.96\textwidth]{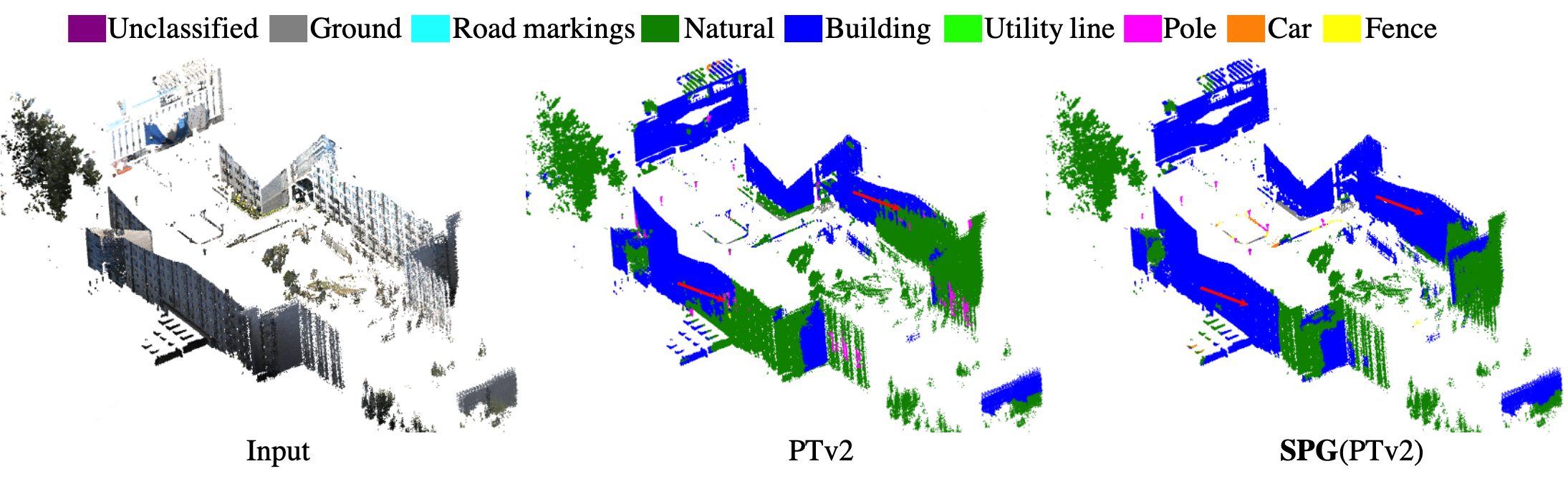}
\caption{Qualitative results of the real-world data. The main part of the point cloud consists of buildings, with some trees surrounding them.} 
\label{fig11}
\end{figure}

\begin{table}[ht] \scriptsize
\centering
\caption{The performance of different balancing methods on S3DIS Area5.}
\label{tab_balance}
\begin{tabular}{c|cccccc>{\columncolor[gray]{0.8}}c}
\toprule
Method &Base  & Focal\cite{lin2017focal} & Weighted CE\cite{aurelio2019learning}  &BNN\cite{zhou2020bbn}  &Meta-Softmax\cite{ren2020balanced}  &NORCAL\cite{pan2021model}  & \textbf{SPG}  \\ \midrule
PTv1\cite{zhao2021point}&70.4  &\makecell{69.2\\($\downarrow$1.2)}  &\makecell{69.6\\($\downarrow$0.8)}  &\makecell{69.3\\($\downarrow$1.1)}  &\makecell{69.8\\($\downarrow$0.6)}  &\makecell{69.9\\($\downarrow$0.5)} &\makecell{\textbf{71.5}\\($\uparrow$1.1)} \\
PTv2\cite{wu2022point} &71.6  &\makecell{70.7\\($\downarrow$0.9)}   &\makecell{70.9\\($\downarrow$0.7)}   & - & - & -&\makecell{\textbf{73.3}\\($\uparrow$1.7)}  \\ 
\bottomrule
\end{tabular}
\end{table}

\noindent
\begin{minipage}{\textwidth}
    \begin{minipage}[h]{0.42\textwidth}
    \scriptsize
    \centering
    \begin{tabular}{ccccc|c}
    \toprule
    $Separate $ & $L_{con}$ & $L_{l1}$ & $L_{l1}^{\prime}$ & $L_{ce}^{\prime}$ & mIoU \\ \midrule
    \rowcolor[gray]{0.8} \checkmark &  \checkmark &  \checkmark      &  \checkmark       &   \checkmark      & \textbf{73.3}    \\
    &  \checkmark &  \checkmark      &  \checkmark       &   \checkmark      &  72.2($\downarrow$1.1)    \\
    \checkmark &          &  \checkmark      &  \checkmark       &   \checkmark      &  72.6($\downarrow$0.7)    \\
    \checkmark & \checkmark          &        &  \checkmark       &   \checkmark      &  72.8($\downarrow$0.5)    \\
    \checkmark & \checkmark          &  \checkmark      &         &   \checkmark      &  71.6($\downarrow$1.7)    \\ \bottomrule
    \end{tabular}
    \makeatletter\def\@captype{table}\makeatother\caption{Ablation study of \textbf{SPG}(PTv2) on S3DIS Area5. "\textit{Separate}" refers to whether to use the separate subspaces in the auxiliary branch.}
    \label{tab4}  
    \end{minipage}
        	\begin{minipage}[h]{0.57\textwidth}
        	\centering
        \includegraphics[height=0.69\textwidth]{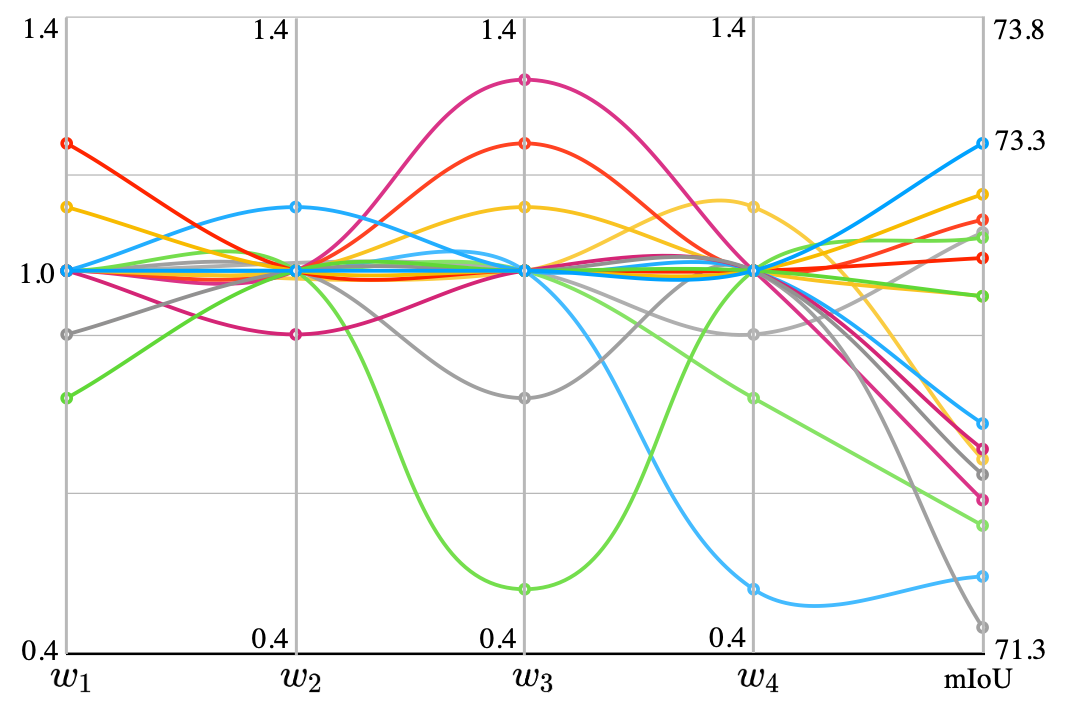}
        \makeatletter\def\@captype{figure}\makeatother\caption{Loss weight adjustment in \cref{equ11} on S3DIS Area5.}
        \label{fig10}
        \end{minipage}
\end{minipage}

\begin{table}[ht]\scriptsize
    \caption{Hyperparameter tuning on S3DIS Area5. $\mathfrak{t}$ represents the number of iterations in each training epoch.} 
    \label{tab3}
    \begin{subtable}{0.5\linewidth}
      \centering
        \subcaption{$\tau$ denotes the temperature in \cref{equ3}.}
        \begin{tabular}{c|ccc >{\columncolor[gray]{0.8}}ccc}
        \toprule
        $\tau$ & 0.02 & 0.05 & 0.06 & \textbf{0.07} & 0.08 & 0.10 \\ \midrule
        mIoU &72.2  &73.0 &72.9 &\textbf{73.3} &72.9 &72.0 \\ \bottomrule
        \end{tabular}
    \end{subtable}
    \begin{subtable}{0.5\linewidth}
      \centering
        \subcaption{$\alpha$ is the smoothing factor in \cref{equ5}.}
        \begin{tabular}{c|ccc >{\columncolor[gray]{0.8}}cccc}
        \toprule
        $\alpha$ & 0.9 & 0.99 & 0.999 & \textbf{1-1/$\mathfrak{t}$} & 1-1/$\frac{\mathfrak{t}}{2}$ & 1-1/$\frac{\mathfrak{t}}{10}$ \\ \midrule
        mIoU  &72.1  &72.6  &73.0 & \textbf{73.3} & 72.8 & 72.3\\ \bottomrule
        \end{tabular}
    \end{subtable} 
\end{table}

\subsection{Qualitative Results}
	\cref{fig8} displays the improvements on the segmentation performance of PTv1 and PTv2 due to the training guidance from \textbf{SPG}, and the corresponding feature distribution maps. In scene (a), due to the proximity of "bookcase" and "clutter" category in the feature spaces of PTv1 and PTv2 (indicated within the grass green and black rectangle), the "bookcase" points (indicated by red arrows) are misclassified as "clutter" points. For the same reason, "table" points are also misclassified as "clutter" points. After training with \textbf{SPG} guidance, the features of "bookcase" and "table" categories become more discriminative, significantly reducing the probability of misclassification. In scene (b), the networks trained without \textbf{SPG} guidance exhibit limited discrimination for the "bookcase" and "clutter" categories in the feature space. This reduced cohesion makes it more susceptible to be misclassified. In the scene shown in \cref{fig9}, which is from ScanNet v2, it is also evident that \textbf{SPG} can assist the segmentation network in achieving improved segmentation results.
	
	We test the models that are pretrained on Toronto-3D on the real-world collected data, and the qualitative results are shown in \cref{fig11}. It can be observed that \textbf{SPG} demonstrates significant advantages in handling real-world data. By assisting the base segmentation network in reducing intra-class feature distances and increasing inter-class distances, \textbf{SPG} markedly reduces the probability of misclassifying point clouds belonging to the "Building" category as "Natural" and "Pole" categories.

\subsection{Ablation and Hyperparameters}
	Point cloud semantic segmentation lacks clear sample statistics (\eg, images in classification, boxes in object detection), making methods\cite{aurelio2019learning, ren2020balanced, pan2021model} that depend on statistical priors challenging to work. Additionally, resampling data\cite{zhou2020bbn, ren2020balanced} will disrupt the global semantics and context information of the input point cloud and re-weighting per-pixel/point loss\cite{lin2017focal, aurelio2019learning} may cause gradient issues during optimization. The proposed method, \textbf{SPG}, does not rely on statistical priors, resampling samples, or point-wise loss adjustments, thus avoiding potential performance degradation of other methods in \cref{tab_balance}. From the ablation experiments in \cref{tab4}, it can be observed that prototypes extracted from separate feature subspaces can more effectively guide the training of segmentation networks, resulting in superior segmentation performance. The loss constraints of each component in \cref{equ11} are beneficial to the improvement of performance. 
	 
	 As shown in \cref{fig10}, when $w1$, $w2$, $w3$, and $w4$ are all set to their default value of $1$, \textbf{SPG} (PTv2) achieves the best performance. This suggests that the losses of each part are balanced without the need for complex adjustments. In \cref{tab3}, \textbf{SPG}(PTv2) achieves the best performance at the temperature $\tau$=0.07 and the smoothing factor $\alpha$=1-1/$\mathfrak{t}$. The above hyperparameter settings can be transferred to different base networks (\eg, PointNet++, PTv1) to achieve the segmentation performance shown in \cref{tab1} on various datasets.

\section{Conclusion}
	This study attempts to address the issue of class imbalance in point cloud semantic segmentation from the perspective of point cloud feature prototypes and proposes a targeted solution called \textbf{SPG}. \textbf{SPG} extracts prototypes from separate feature subspaces to guide the training of segmentation network. This process makes the distribution of homogeneous features more similar and the distribution of heterogeneous features more different, which effectively mitigates the cognitive bias introduced by class imbalance. Extensive experiments conducted on different networks and datasets confirm the effectiveness of \textbf{SPG} in enhancing the discrimination of minority category features, ultimately improving overall segmentation accuracy.

\section*{Acknowledgements}
It is sponsored by State Key Laboratory of Intelligent Green Vehicle and Mobility under Project No. KFY2416. Thanks to Prof. Wenguang Wang of Beihang University for providing the collected real-world data.

\bibliographystyle{splncs04}
\bibliography{spg}

\end{document}